# Probabilistic Conflict Resolution in Hierarchical Hypothesis Spaces

Tod S. Levitt
Advanced Information & Decision Systems
201 San Antonio Circle, Suite 286
Mountain View, California 94040

## Introduction

Artificial intelligence applications such as industrial robotics, military surveillance, and hazardous environment clean-up, require situation understanding based on partial, uncertain, and ambiguous or erroneous evidence. It is necessary to evaluate the relative likelihood of multiple possible hypotheses of the (current) situation faced by the decision making program. Often, the evidence and hypotheses are hierarchical in nature. In image understanding tasks, for example, evidence begins with raw imagery, from which ambiguous features are extracted which have multiple possible aggregations providing evidential support for the presence of multiple hypotheses of objects and terrain, which in turn aggregate in multiple ways to provide partial evidence for different interpretations of the ambient scene. Information fusion for military situation understanding has a similar evidence/hypothesis hierarchy from multiple sensor through message level interpretations, and also provides evidence at multiple levels of the doctrinal hierarchy of military forces.

Real world hierarchical evidential reasoning for situation understanding differs from classical confirmation or denial of alternative hypotheses in that the multiple hypotheses of a situation interpretation cannot be specified a priori. For instance, in image understanding systems, there typically is a finite dictionary of modeled objects, but there are no models of scenes in which these objects will appear. Instead there is a representation of which object models and relations are consistent or inconsistent with other objects.

It follows that the hypotheses for which we have prior models, are not necessarily exclusive. The global scene interpretation consists of some consistent set of hypotheses, but in general, we have no prior model of the global situations. Furthermore, even if we have a numerical scheme for accruing the weight of evidence supporting any given object (model) hypothesis, we still have no obvious method for deciding which set of hypothesized objects are the best global interpretation of the scene. The following example shows that the strategy of constructing a global interpretation by taking the hypothesis with strongest support, eliminating from further consideration all hypotheses inconsistent with this best supported hypothesis, and iterating the process to obtain a global interpretation, may yield very different results from a strategy which directly reasons about global interpretations.

Figure 1a shows the logical sharing of evidence by two hypotheses $H_1$, and $H_2$. Figure 1b shows the creation of new hypotheses $H_3$ and $H_4$, to explicitly represent the alternatives to the conflict. $H_3$ is hypothesis $H_1$, but without claiming support from $h_2$, while $H_4$ is $H_2$, but without claiming $h_2$. We now see that the maximal sets of consistent situation interpretations are represented by $\{H_1,H_4\}$ or $\{H_2,H_3\}$ or $\{H_3,H_4\}$ (in this last, no hypothesis claims $h_2$, i.e., $h_2$ is being interpreted as unassociated or a false alarm). Explicit representation of conflict alternatives in the hypothesis space makes it easy to write fast algorithms to extract globally consistent sets of hypotheses.

Suppose the probabilities, p, for the hypotheses of Figure 1b are $p(H_1) = .52$, $p(H_2) = .19$, $p(H_3) = .23$, and $p(H_4) = .06$. This corresponds to evidence $h_1$ (.7) being relatively certain and $h_2$ (.1) and $h_3$ (.2) relatively uncertain. Then following a "strongest hypothesis first" strategy, we obtain the global interpretation of $\{H_1, H_4\}$. The accrual scheme used here is presented in the next section of this paper.

Consider, alternatively, the set of possible global interpretations. For Figure 1b this is the list: $A = \{(H_1, H_4), (H_2, H_3), (H_3, H_4)\}$. We assume that each hypothesis has a certainty value attached which was obtained by numerically accruing evidence for each hypothesis individually, ignoring the other hypotheses. Our objective is to apply numerical accrual techniques to rank order the list of consistent interpretations. The approach suggested here is to make the list $A$ into a probability space.

We use the standard mathematical techniques for re-norming $A$. Some task-dependent assumptions may have to be made. We must first derive a probability for the "joint events" $(H_1, H_4)$ etc. If we assume that (e.g.) $H_1$ and $H_4$ are independent, then the probability of $(H_1, H_4)$ is simply the product of the probabilities of $H_1$ and $H_4$. This seems a reasonable assumption, since if $H_1$ and $H_4$ are mutually consistent, then they are (probably) supported by disjoint bodies of (non-contradictory) evidence. We now have a certainty value for each member of $A$. We make $A$ into a probability space by normalization. The list $A$ is now rank ordered by these probabilities.

Then, following the above program, we have, after normalization, $p(H_1, H_4) = .36$, $p(H_2, H_3) = .49$, $p(H_3, H_4) = .15$. So, although $H_1$ is the highest probability hypothesis individually, it is not part of the strongest global situation interpretation.

In the following, we formalize this method of conflict resolution for dynamic determination of best global interpretation in a non-exclusive, model-based hypothesis space. We begin with a technique for utilizing our prior models of (non-exclusive) hypotheses to aid us in numerically accruing evidence for hypothesis hierarchies. This is followed by a scheme for conflict resolution for situation understanding. Conflict resolution at the $N$-th level immediately yields a best-hypothesis-tree extraction for the entire hierarchy. Using these techniques, we show that in the case where all situation interpretations at level $m$, i.e., consistent hypothesis sets, are equal length, and $\{h_1, \ldots, h_n\}$ is selected at level $m$ by propagating the best hypothesis interpretation downward for level $m+1$, there is a strictly less than

$$\left( \left[ \sum_{i=1}^{n} p(h_i) \right]^n - \prod_{i=1}^{n} p(h_i) \right) / n!$$

chance of randomly extracting a less than optimal interpretation at level $m$.

## Numerical Support for Hypothesis Hierarchies

We assume a hierarchical hypothesis space of $N$ levels, where the hypotheses at level $m$ provide evidential support for those at level $m+1$. Each pair of hypotheses at level $m$ are either mutually consistent or are in conflict. A non-conflicted hypothesis hierarchy is pictured in Figure 2. Hypotheses must be generated and associated according to pattern matching routines (or rules, etc.) based on the model of a level $m+1$ hypothesis in terms of observed or inferred (evidence) hypotheses at level $m$. Level $N$ (top level) hypotheses represent the largest grain-sized objects for which prior models exist.

Partial, ambiguous and false evidence can cause generation of hypotheses which conflict with existing hypotheses. For example, if an image feature edge is claimed as support by two different tool hypotheses, then both cannot be true (solely) on the basis of that evidence. In general, incompatible sharing of evidence or component hypotheses is one generic type of

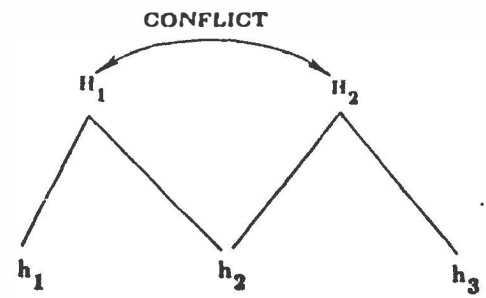

a) Logical conflict representation

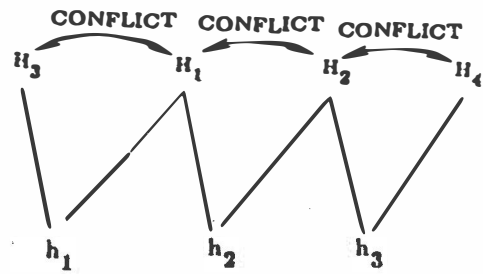

b) Explicit representation of conflict alternatives

Figure 1: Representation of Hypothesis Conflicts

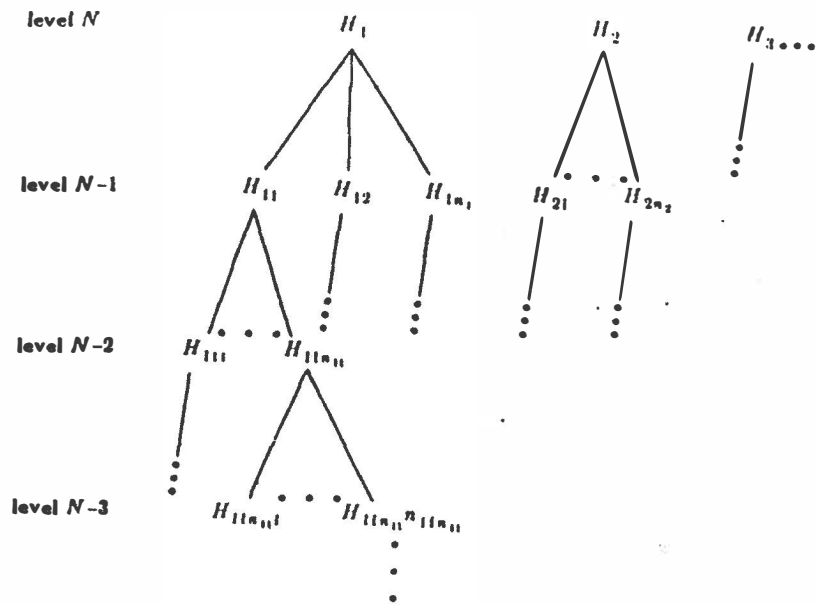

Figure 2: A Consistent Hypothesis Space

hypothesis conflict. However, hypotheses may conflict without sharing components. In a legal reasoning example, hypotheses which place the defendant at different locations at the same time may be supported by disjoint sets of evidence, but are, nonetheless, in conflict. Similarly, the hypotheses of presence of two military forces maybe supported by individual radio intercepts, but be in conflict because the forces are located too close together relative to military doctrine. In general, conflicts caused by other than shared evidence or supporting hypotheses are dependent on the particular situations represented in the hypothesis space.

Conflicts should be searched for at the time of hypothesis generation and/or associations. Once recognized, all consistent interpretations of the hypothesis space should be computed and explicitly represented in the hypothesis space. This principle is illustrated in Figure 1.

Given a hierarchical hypothesis space with conflicts, we invoke the rule that all conflicts must be propagated upward. Keep in mind that this increases the generation of alternative hypotheses which do not claim conflicting evidence.

After upward propagation of conflicts, at the top level of the hypothesis hierarchy we have a set of hypotheses with conflict pointers as illustrated in Figure 3a. Consistent interpretations of the situation represented by these hypotheses amounts to extracting maximal sets of non-conflicting hypotheses. The explicit representation of alternatives and conflicts makes this process realizable in a simple recursive algorithm, as in [Levitt - 85]. The results of this process for Figure 3a are shown in Figure 3b.

We require that the following rules be followed in hypothesis generation.

1) (Conflict definition)
   If two level $m+1$ hypotheses are associated to a common supporting sub-hypothesis at level $m$, then this is a conflict which must be explicitly declared.

2) (Generation of alternatives)
   In the case of conflict through common evidence association, the alternative, non-conflicting hypotheses must be generated. That is, if $H_1$ is associated to (i.e., supported by) a set of hypotheses $S_1$, and $H_2$ is associated to a set $S_2$, and $S_1$ intersect $S_2$ is the set $S_{12}$, then $H_3$ must be generated and associated to $S_1-S_{12}$ and $H_4$ must be generated and associated to the hypothesis set $S_2-S_{12}$.

3) (Uniqueness of association)
   No two level $m+1$ hypotheses may be associated to identical sub-hypothesis sets at level $m$.

Together these three rules imply that for any connected hypothesis hierarchy, all the hypotheses at level $m$ are associated to the same number of hypotheses at level $m+1$. (This value may vary between levels.)

We assume that probability values come in at the lowest level of the hierarchy. The initial problem is to propagate these probabilities upward. That is, given a hypothesis, $H$, supported by (evidence) sub-hypotheses $\{h_1, \ldots, h_n\}$, we wish to compute $p(H \mid h_1, \ldots, h_n)$. We assume $H$ is at the $m+1$ level of the hierarchy, and that the values $p(h_i)$ have already been accrued at the $m$-th level. $H$ has been associated to the $h_i$ based on a prior model of $H$ which requires $n+k$ component hypotheses, of which $n$ have been observed, namely the set $\{h_1, \ldots, h_n\}$.

Let $a_m$ be the number of $m+1$ level hypotheses associated to each level $m$ hypothesis. (As noted earlier, this value is constant at level $m$.) We define

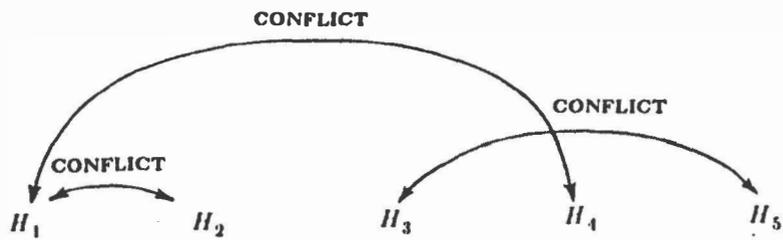

a) Hypothesis conflicts at the top level

$$\{ \begin{array}{l} (H_1, H_3) \\ (H_1, H_5) \\ (H_2, H_3, H_4) \\ (H_2, H_4, H_5) \end{array} \}$$

b) Maximal consistent global sets of hypotheses

Figure 3: Hypothesis Conflicts and Consistent Interpretations

$$p(H \mid h_1, \ldots, h_n) = \left(\frac{n}{n+k}\right)\left(\frac{1}{a_m}\right)\sum_{i=1}^{n} p(h_i)$$

We claim that this is a probability function on any set of level $m+1$ hypotheses supported by full evidence at level $m$. To see this, suppose a maximal consistent set of level $m$ hypotheses is $\{h_1, \ldots, h_r\}$, $\sum p(h_r) = 1$. Let $\{H_1, \ldots, H_s\}$ be the $m+1$ level hypotheses supported by these. Then for this space $\frac{n}{n+k} = 1$. Further, since the level $m$ hypotheses are not in conflict, it is not unreasonable to assume that

$$p(H_i \mid h_1, \ldots, h_r) = p(H_i \mid h_{i_1}, \ldots, h_{i_n})$$

where $h_{i_1}, \ldots, h_{i_n}$ are the hypotheses associated to $H_i$.

It follows that

$$\sum_{i=1}^{s} p(H_i \mid h_1, \ldots, h_r)$$
$$= \sum_{i=1}^{s} p(H_i \mid h_{i_1}, \ldots, h_{i_n})$$
$$= \frac{1}{a_m} \sum_{i=1}^{s} \sum_{j=i_1}^{i_n} p(h_j)$$

But each $p(h_j)$ occurs exactly $a_m$ times in this sum, so

$$= \frac{1}{a_m}\left(a_m \sum_{j=1}^{r} p(h_j)\right)$$

$$= 1.$$

Referring to Figure 1, let $p(h_1) = .7$, $p(h_2) = .1$ and $p(h_3) = .2$, then $a_m = 2$. Assume without loss of generality that all the $H_i$ have precisely two components required in their model. Then we compute

$$p(H_1 \mid h_1, h_2) = \left(\frac{2}{2}\right)\left(\frac{1}{2}\right)(.7+.1) = .400$$

$$p(H_2 \mid h_2, h_3) = \left(\frac{2}{2}\right)\left(\frac{1}{2}\right)(.1+.2) = .150$$

$$p(H_3 \mid h_1) = \left(\frac{1}{2}\right)\left(\frac{1}{2}\right)(.7) = .175$$

$$p(H_4 \mid h_3) = \left(\frac{1}{2}\right)\left(\frac{1}{2}\right)(.2) = .050$$

Normalizing, we obtain the values given in the introduction. Notice that this accrual scheme has the pleasant advantage that associating additional supporting evidence increases the probability of a hypothesis.

## Probabilistic Conflict Resolution

From the set of $N$ th level (conflicting) hypotheses, we can construct a set, $A$, of mutually exclusive and exhaustive interpretations of the global situation using a straightforward recursive algorithm. We now make $A$ a probability space in the following manner.

A typical element of $A$ is a maximal set $S = \{H_1, \ldots, H_n\}$ of non-conflicting hypotheses. It seems reasonable to assume that they are all mutually independent, since they are based on our top level models and depend on disjoint bodies of evidence. Thus, if we let $E = \{E_1, \ldots, E_n\}$, where $E_i$ is the evidence supporting $H_i$ (i.e., $E_i = \{h_{i_1}, \ldots, h_{i_{n_i}}\}$) we have:

$$p(S \mid E) = p(H_1, \ldots, H_n \mid E) = \prod_{i=1}^{n} p(H_i \mid E) = \prod_{i=1}^{n} p(H_i \mid E_i)$$

The $p(H_i \mid E_i)$ terms are calculated as in the previous section. Since all the $S$ in $A$ are mutually exclusive and exhaustive, we can create a probability space by dividing $p(S \mid E)$ by a normalization factor, $K$, calculated as

$$K = \sum_{S \in A} P(S \mid E)$$

To extract the best global interpretation of the situation represented in the hierarchy, we claim it is sufficient to take the maximal consistent interpretation given by the set $S$ with greatest probability $p(S \mid E)/K$, and extract the unique consistent hierarchical subspace associated below $S$ for a hypothesis-tree. In the following, we show that this is a "probably best" hypothesis tree in the sense that the probability that the set of hypotheses selected at level $m$, $(\{h_1, \ldots, h_n\})$, by propagating the global interpretation downward from level $m+1$, will not be the best global interpretation at level $m$ is less than

$$\frac{\left[\sum_{i=1}^{n} p(h_i)\right]^n - \prod_{i=1}^{n} p(h_i)}{n!}.$$

Proof is only given for the case when all interpretations at level $m$ have the same number of hypotheses in them.

Suppose that the set of hypotheses $\{h_1, \ldots, h_n\}$ are the level $m$ support for the best global interpretation of hypotheses at level $m+1$. Then $\{h_1, \ldots, h_n\}$ were essentially selected by comparing a product of sums of probabilities of subsets of $\{h_1, \ldots, h_n\}$ against those of other (non-conflicted) hypothesis sets. It is possible that if we extract the best global interpretation at level $m$ by our conflict resolution scheme, i.e., by taking products of probabilities, that the set $\{h_1, \ldots, h_n\}$ may not have greatest probability. For example, if $n=2$ and $p(h_1) = .42$, $p(h_2) = .11$, $p(h_3) = .21$, $p(h_4) = .26$, then $p(h_1) + p(h_2) = .537 > .47 = p(h_3) + p(h_4)$ but $p(h_1) * p(h_2) = .046 < .055 = p(h_3) * p(h_4)$. However, the random chance of this occurring is not great, and it rapidly decreases as $n$ increases.

To see this note first that $\prod_i \left(\sum_j p(h_{ij})\right) \leq \sum_{i,j} p(h_{ij})$ for any partitioning $\{h_{11}, \ldots, h_{1n_1}, \ldots, h_{k1}, \ldots, h_{kn_k}\}$ of $\{h_1, \ldots, h_n\}$, since all the $p(h_{ij})$ are between zero and one. For the sake of our approximation, we are interested in the space of probabilities whose sum is less than the sum of our extracted hypothesis but whose product is greater.

Now suppose $\sum_{i=1}^n p(h_i) = S$ and $\prod_{i=1}^n p(h_i) = P$. The space of probabilities fulfilling each of these criterion are two surfaces (a hyperplane and hyperboloid respectively) in Euclidean $n$-space. Because all the $p(h_i)$ are between zero and one, we know $P < S$. Figure 4 illustrates the situation for $n=2$. Here, the shaded region indicates those sets of $n$ probabilities whose sum will be less than $S$, but whose product will be greater than $P$.

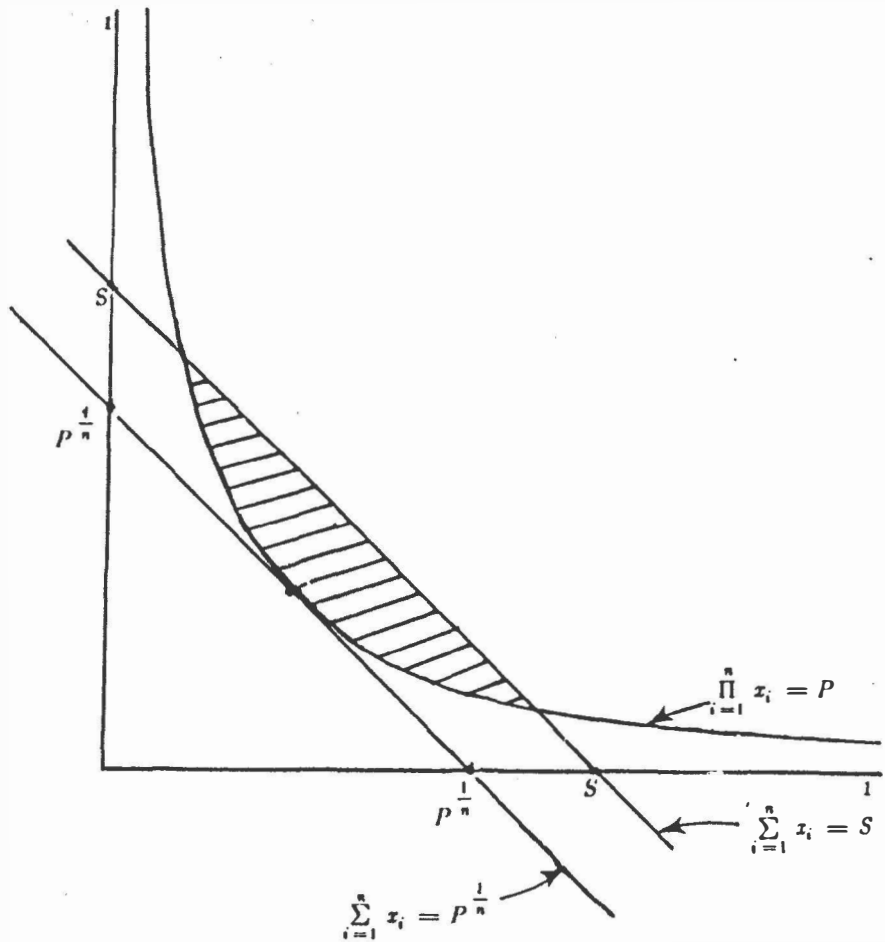

Figure 4: Volume of Probabilities Better Than $P$ but Less Than $S$

We can view the volume of the shaded region (in $n$ dimensions) as the random probability that there is a set of $m$-level hypotheses which will have greater product probability than that selected by the conflict resolution scheme. (Note that the volume is a subset of the unit cube in $n$-space.)

The volume is bounded above by the volume between the hyperplanes $\sum_{i=1}^{n} x_i = S$ and $\sum_{i=1}^{n} x_i = P^{\frac{1}{n}}$ in the unit cube. We can calculate this volume as

$$\int_0^S \int_0^{x_{n-1}} \cdots \int_0^{x_2} x_1 \, dx_1 \cdots dx_{n-1} - \int_0^{P^{\frac{1}{n}}} \int_0^{x_{n-1}} \cdots \int_0^{x_2} x_1 \, dx_1 \ldots dx_{n-1}$$

which is
$$\frac{S^n - P}{n!} = \frac{\left[\sum_{i=1}^{n} p(h_i)\right]^n - \prod_{i=1}^{n} p(h_i)}{n!}.$$

In general, this quantity will be maximized when $S=1$ and $P$ is minimized, which occurs when $p(h_1) = p(h_2) = \ldots = p(h_n) = \frac{1}{n}$. We see that

$$0 \leq \lim_{n \to \infty} \frac{S^n - P}{n!} \leq \lim_{n \to \infty} \frac{1 - (\frac{1}{n})^n}{n!} = 0.$$

(Notice that $S^n - P \geq 0$ because the arithmetic mean of $n$ variables is greater than the geometric mean of $n$ variables, see [Mays, 1983].)

In fact, the random chance of having a less than optimal hypothesis tree extraction shrinks rapidly as $n$ increases. For $n=2$, the chances are less than .375, for $n=3$, less than .16, for $n=4$, less than .04 and for $m=5$, less than .008.